\documentclass[sigconf]{acmart}

\usepackage{algorithm}
\usepackage{algorithmic}
\usepackage{threeparttable}
\usepackage{amsmath}
\usepackage{subfigure}
\usepackage{newfloat}
\usepackage{listings}
\usepackage{multirow}
\usepackage{multicol}
\usepackage{bm}
\usepackage{graphicx}
\usepackage{subfigure}
\AtBeginDocument{%
  \providecommand\BibTeX{{%
    \normalfont B\kern-0.5em{\scshape i\kern-0.25em b}\kern-0.8em\TeX}}}

\setcopyright{acmcopyright}
\copyrightyear{2025}
\acmYear{2025}
\acmDOI{XXXXXXX.XXXXXXX}

\acmConference[Conference acronym 'XX]{Make sure to enter the correct
  conference title from your rights confirmation emai}{April 29,
  2025}{Sydney, Australia}

\acmPrice{15.00}
\acmISBN{978-1-4503-XXXX-X/18/06}

\begin{document}

\title{Learning Structure-enhanced Temporal Point Processes with Gromov-Wasserstein Regularization}

% \author{
%     %Authors
%     % All authors must be in the same font size and format.
%     Qingmei Wang\textsuperscript{\rm 1}, Fanmeng Wang\textsuperscript{\rm 1}, Bing Su\textsuperscript{\rm 1,3}, Hongteng Xu\textsuperscript{\rm 1,3}\thanks{Corresponding author: hongtengxu@ruc.edu.cn}
% }
% \affiliations{
%     %Afiliations
%     \textsuperscript{\rm 1}Gaoling School of Artificial Intelligence, Renmin University of China\\
%     \textsuperscript{\rm 3}Beijing Key Laboratory of Big Data Management and Analysis Methods\\
   
% }

% 作者 1
\author{Qingmei Wang}
\affiliation{%
  \institution{Gaoling School of Artificial Intelligence}
  \city{}
  \country{Renmin University of China}
}
\email{}

% 作者 2
\author{Fanmeng Wang}
\affiliation{%
  \institution{Gaoling School of Artificial Intelligence}
  \city{}
  \country{Renmin University of China}
}
\email{}

% 作者 3
\author{Bing Su}
\affiliation{%
  \institution{Gaoling School of Artificial Intelligence\\Beijing Key Laboratory of Big Data Management and Analysis Methods}
  \city{}
  \country{Renmin University of China}
}
\email{}

% 作者 4
\author{Hongteng Xu}
\affiliation{%
  \institution{Gaoling School of Artificial Intelligence\\Beijing Key Laboratory of Big Data Management and Analysis Methods}
  \city{}
  \country{Renmin University of China}
}
\email{hongtengxu@ruc.edu.cn}

% \author{Qingmei Wang}
% \affiliation{%
%   \institution{Gaoling School of Artificial Intelligence}
%   \city{ Renmin University of China}
%   \country{}}
% \email{}

% \author{Fanmeng Wang}
% \affiliation{%
%   \institution{Gaoling School of Artificial Intelligence}
%   \city{Renmin University of China}
%   \country{}}
% \email{}

% \author{Bing Su}
% \affiliation{%
%   \institution{Gaoling School of Artificial Intelligence\\Beijing Key Laboratory of Big Data Management and Analysis Methods}
%   \city{}
%   \country{Renmin University of China}}
% \email{}

% \author{Hongteng Xu}
% \email{qingmeiwang@ruc.edu.cn}
% \orcid{1234-5678-9012}
% \author{G.K.M. Tobin}
% \authornotemark[1]
% \email{webmaster@marysville-ohio.com}
% \affiliation{%
%   \institution{Institute for Clarity in Documentation}
%   \streetaddress{P.O. Box 1212}
%   \city{Dublin}
%   \state{Ohio}
%   \country{USA}
%   \postcode{43017-6221}
% }

%%
%% By default, the full list of authors will be used in the page
%% headers. Often, this list is too long, and will overlap
%% other information printed in the page headers. This command allows
%% the author to define a more concise list
%% of authors' names for this purpose.
\renewcommand{\shortauthors}{Qingmei Wang, et al.}

\begin{abstract}
Real-world event sequences are often generated by different temporal point processes (TPPs) and thus have clustering structures. 
Nonetheless, in the modeling and prediction of event sequences, most existing TPPs ignore the inherent clustering structures of the event sequences, leading to the models with unsatisfactory interpretability. 
In this study, we learn structure-enhanced TPPs with the help of Gromov-Wasserstein (GW) regularization, which imposes clustering structures on the sequence-level embeddings of the TPPs in the maximum likelihood estimation framework.
% In this study, we propose a novel framework that incorporates Optimal Transport (OT) theory, with a particular focus on the Gromov-Wasserstein (GW) distance, into scalable nonparametric regularization.
% In principle, our method trains a single TPP, imposing clustering structures on the embeddings of its event sequences. 
In the training phase, the proposed method leverages a nonparametric TPP kernel to regularize the similarity matrix derived based on the sequence embeddings. 
% Leveraging the nonparametric TPP clustering method, we design a kernel matrix to regularize the similarity matrix of the sequence embeddings during training. 
In large-scale applications, we sample the kernel matrix and implement the regularization as a Gromov-Wasserstein (GW) discrepancy term, which achieves a trade-off between regularity and computational efficiency.
The TPPs learned through this method result in clustered sequence embeddings and demonstrate competitive predictive and clustering performance, significantly improving the model interpretability without compromising prediction accuracy.
\end{abstract}

%%
%% The code below is generated by the tool at http://dl.acm.org/ccs.cfm.
%% Please copy and paste the code instead of the example below.
%%


%%
%% Keywords. The author(s) should pick words that accurately describe
%% the work being presented. Separate the keywords with commas.

\maketitle

\section{Introduction}
Temporal point processes (TPPs) are powerful tools for modeling events that occur sequentially in continuous-time domain~\cite{kingman1992poisson,ross1996stochastic}.
They have achieved encouraging performance in many applications, e.g., healthcare data analysis~\cite{enguehard2020neural}, social network modeling~\cite{zhou2013learning,li2014learning,zhao2015seismic}, financial data analysis~\cite{linderman2014discovering,bacry2015hawkes} and web science~\cite{yao2021stimuli,kong2023interval,junuthula2019block}. 
Furthermore, in web science, TPPs are particularly useful for modeling and understanding the temporal dynamics of online events, such as user interactions~\cite{junuthula2019block,hatt2020early}, content generation~\cite{bao2015modeling}, and information diffusion~\cite{farajtabar2015coevolve,nickel2021modeling}. 
Despite their usefulness, the above TPPs seldom consider the clustering structures hidden in event sequences. 
In fact, real-world event sequences often yield different generative mechanisms and thus belong to different clusters.
For example, patients suffering from different diseases often have different admission behaviors. 
Laborers in different industries have various career advancement trajectories and job-hopping experiences. 
Ignoring such clustering structures may lead to the model misspecification issue, doing harm to the interpretability and prediction power of the models.

% Optimal Transport (OT) provides a promising solution for addressing these challenges. 
% OT techniques, such as the Gromov-Wasserstein (GW) distance, are effective for comparing metric spaces, making them ideal for capturing clustering structures within event embeddings~\cite{chhoa2024metric}.
% By aligning event sequences in latent space, GW-based regularization enhances the interpretability and clustering quality of sequence embeddings without compromising prediction accuracy~\cite{gong2022gromov,chowdhury2021generalized}. 
% Furthermore, GW distance’s computational efficiency via approximation techniques makes it suitable for large-scale datasets~\cite{xu2019scalable}.

\begin{figure}[t]
    \centering
    \includegraphics[height=8cm]{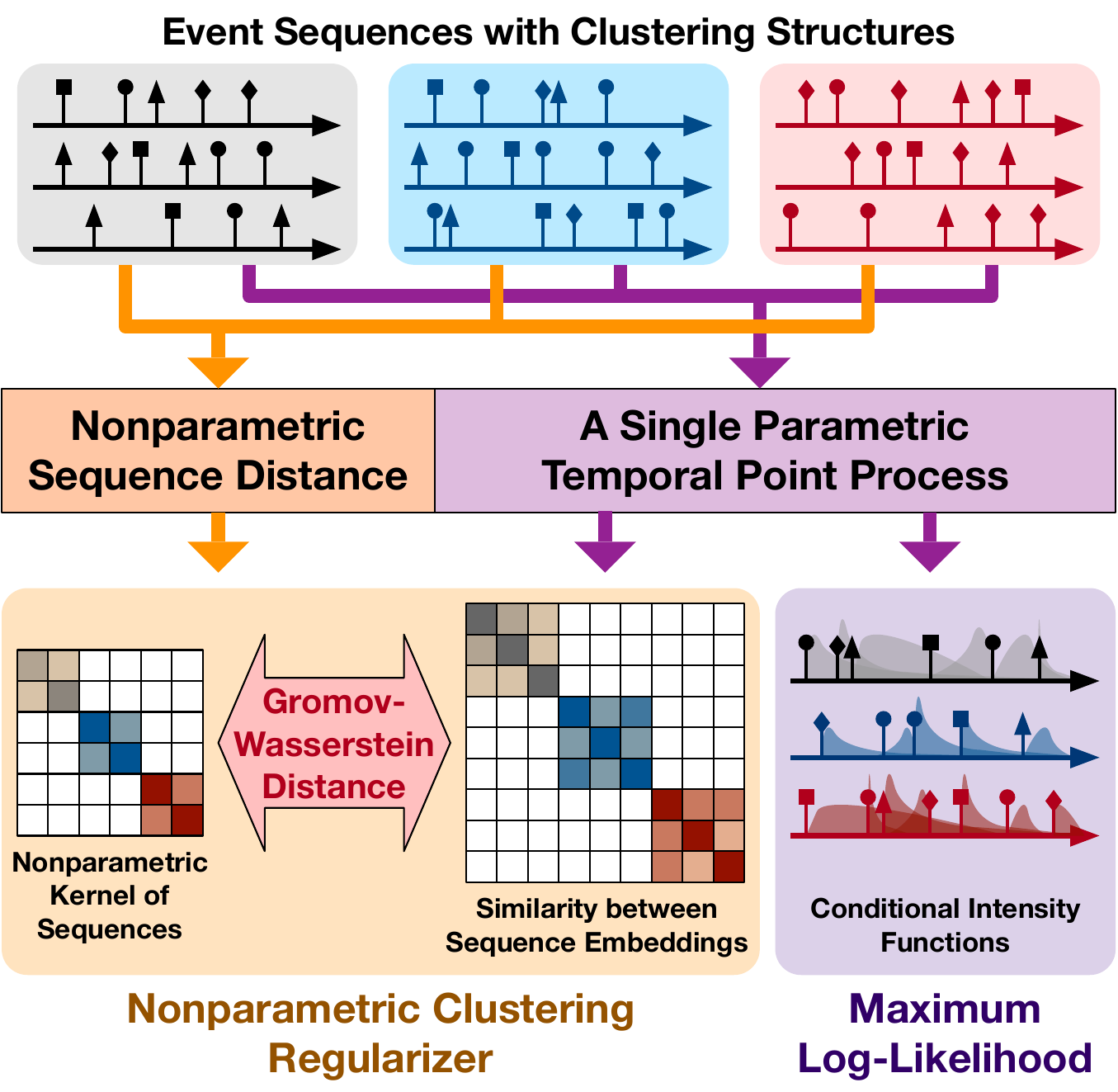}
    \caption{The scheme of the proposed method.}
    \label{fig:scheme}
\end{figure}

To learn TPPs with both predictive and clustering capabilities, in this study, we propose a novel regularizer with the help of the Gromov-Wasserstein discrepancy~\cite{memoli2011gromov}, which learns structure-enhanced TPPs effectively in the framework of maximum likelihood estimation.
As illustrated in Figure~\ref{fig:scheme}, our method learns a single parametric TPP and imposes clustering structures on the embeddings of different event sequences based on a nonparametric clustering regularizer. 
In particular, leveraging the nonparametric clustering method in~\cite{iwayama2017definition}, we design a kernel matrix to regularize the similarity matrix of the sequence embeddings. 
Plugging the regularizer into the maximum likelihood estimation (MLE) framework, we learn the TPP with nonparametric clustering guidance.
To make the proposed regularizer applicable for large-scale applications, we construct a small kernel matrix from a subset of event sequences and implement the regularizer as a Gromov-Wasserstein discrepancy term~\cite{memoli2011gromov}.
As a result, the structure-enhanced TPP can be learned by stochastic gradient descent, with low computational complexity.

Essentially, the proposed regularizer is highly flexibility, applying to arbitrary TPPs that can derive sequence embeddings and arbitrary learning paradigms. 
It leads to a scalable and effective solution to both event sequence clustering and prediction, whose complexity is independent of the number of clusters. 
Experiments on synthetic datasets highlight the effectiveness of our method, particularly in combining the strengths of both parametric and nonparametric TPP models.
The TPPs trained with the regularizer achieve competitive predictive accuracy while producing clustered sequence embeddings that significantly enhance model interpretability.

\section{Proposed Method}

\subsection{Event Sequence Embedding and Clustering}

Denote an event sequence with $N$ events as $\bm{s}=\{(t_n, c_n)\}_{n=1}^{N}$, where the tuple $(t_n, c_n)$ is the $n$-th event, $t_n\in [0,T]$ is its timestamp, and $c_n\in\mathcal{C}=\{1,...,C\}$ is its event type. 
A parametric temporal point process (TPP) is often represented as a multivariate counting process, denoted as $\bm{N}(\theta)=\{N_{c}(t;\theta)\}_{c\in\mathcal{C},t\in [0,T]}$, where $\theta$ represents the model parameter and $N_c(t;\theta)$ is a stochastic process counting the number of the type-$c$ events till time $t$.
The TPP captures the dynamics of event sequence by a multivariate conditional intensity function, denoted as $\bm{\lambda}(t;\theta)=\{\lambda_{c}(t;\theta)\}_{c\in\mathcal{C},t\in [0,T]}$, where 
\begin{eqnarray}\label{eq:intensity}
\lambda_{c}(t;\theta)=\frac{\mathrm{d} \mathbb{E}[N_c(t;\theta)|\mathcal{H}_t^{\mathcal{C}}]}{\mathrm{d}t},~\forall c\in\mathcal{C}.
\end{eqnarray}
In~\eqref{eq:intensity}, $\lambda_{c}(t;\theta)$ represents the expected instantaneous rate of the type-$c$ event happening at time $t$ given the historical events $\mathcal{H}_t^{\mathcal{C}}=\{(t_n,c_n)\in\bm{s}|t_n<t\}$.

Given a set of event sequences, i.e., $\mathcal{S}=\{\bm{s}_m\}_{m=1}^{M}$, we often learn a TPP in the following maximum likelihood estimation (MLE) framework~\cite{liniger2009multivariate,zhou2013learning}:
\begin{eqnarray}\label{eq:mle}
\begin{aligned}
    \sideset{}{_{\theta}}\min -\sideset{}{_{m=1}^{M}}\sum\log\mathcal{L}(\bm{s}_m;\theta).
\end{aligned}
\end{eqnarray}
Here,  M is the total number of the set of event sequences and m is the index of the event sequence, $\mathcal{L}(\bm{s}_m;\theta)$ is the likelihood of the sequence $\bm{s}_m=\{(t_{n,m},c_{n,m})\}_{n=1}^{N_m}$, which is formulated based on the conditional intensity function, i.e.,
\begin{eqnarray}\label{eq:nll}
\begin{aligned}
\mathcal{L}(\bm{s};\theta)
=\sideset{}{_{n=1}^{N_m}}\prod\lambda_{c_{n,m}}(t_{n,m})\exp\Bigl(-\sideset{}{_{c\in\mathcal{C}}}\sum\int_{0}^{T}\lambda_c(s)\mathrm{d}s\Bigr).
\end{aligned}
\end{eqnarray}

As shown in~\cite{wang2023hierarchical}, most existing TPPs, especially those based on neural networks, can embed event sequences when calculating their conditional intensity functions. 
The embeddings of $\bm{s}_m$ can be obtained by the aggregation of the event-level embeddings, i.e.,
\begin{eqnarray}\label{eq:emb}
    \bm{h}_m = \text{Pooling}(\bm{h}_{n,m})\in\mathbb{R}^{D},
\end{eqnarray}
where $\text{Pooling}$ represents an arbitrary pooling method.\footnote{In this study, we simply apply the mean-pooling operation.}

For the aforementioned neural TPPs, the event-level embeddings are used to compute the conditional intensity functions, and accordingly, predict future events, while the sequence-level embeddings can be used to measure the similarity among the event sequences. 
In this study, given arbitrary two sequence-level embeddings, we apply a Gaussian kernel to measure their similarity, leading to the following kernel matrix:
\begin{eqnarray}\label{eq:kernel1}
\begin{aligned}
    \bm{K}(\{\bm{h}_m\}_{m=1}^{M})=[\kappa(\bm{h}_m,\bm{h}_{m'})]\in\mathbb{R}^{M\times M},
\end{aligned}
\end{eqnarray}
where $\kappa(\bm{h}_m,\bm{h}_{m'})=\exp(-\frac{\|\bm{h}_m-\bm{h}_{m'}\|_2^2}{2\sigma^2})$ and $\sigma$ is the bandwidth of the kernel function.

When the sequence-level embeddings are learned with discriminative power, we can derive the clustering structure of the event sequences by applying spectral clustering algorithm~\cite{ng2001spectral} to the matrix in~\eqref{eq:kernel1}. 
In this study, we would like to enhance the interpretability of TPP models via imposing clustering structure guidance on their sequence-level embeddings, leading to the following Gromov-Wasserstein regularization strategy.

\subsection{Learning TPPs with Gromov-Wasserstein Regularization} 

\subsubsection{Learning Framework}
Mathematically, given an event sequence $\bm{s}_m=\{(t_{n,m},c_{n,m})\}_{m=1}^{N_m}$, we can represent each event as a $(C+1)$-dimensional ``event vector'', denoted as $\bm{e}_{n,m}=[t_{n,m};\bm{c}_{n,m}]$, where $\bm{c}_{n,m}\in\{0,1\}^C$ is a one-hot vector indicating the event type $c_{n,m}$. 
For each event vector, the maximum value of the first element is $T$, and the maximum value of each remaining element is $1$. 
Let $\{0,1,...,C\}$ be the index set of the event vector and $\mathcal{I}\in\{0,1,...,C\}$ be an index subset. 
For arbitrary two event sequences, i.e., $\bm{s}_m=\{(t_{n,m},c_{n,m})\}_{m=1}^{N_m}$ and $\bm{s}_{m'}=\{(t_{n, m'},c_{n, m'})\}_{m'=1}^{N_{m'}}$, we can represent their events as vectors and measure the difference between the two sequences based on a subset of their event vectors' elements, i.e., 
\begin{eqnarray}\label{eq:diff}
\begin{aligned}
    d_{\mathcal{I}}(\bm{s}_m,\bm{s}_{m'})
    =&\Bigl(\sideset{}{_{n,n'=1}^{N_m}}\sum\sideset{}{_{i\in\mathcal{I}}}\prod(r_{i}-|e_{n,m}^i - e_{n',m}^i|) + \\
    &\sideset{}{_{n,n'=1}^{N_{m'}}}\sum\sideset{}{_{i\in\mathcal{I}}}\prod(r_{i}-|e_{n,m'}^i - e_{n',m'}^i|) - \\
    &2\sideset{}{_{n,n'=1}^{N_m,N_{m'}}}\sum\sideset{}{_{i\in\mathcal{I}}}\prod(r_{i}-|e_{n,m}^i - e_{n',m'}^i|)\Bigr)^{1/2},
\end{aligned}
\end{eqnarray}
where $r_i$ represents the maximum value for the $i$-th element of event vector, and $e_{n,m}^i$ represents the $i$-th element of the event vector $\bm{e}_{n,m}$.

Then, we enumerate all index subsets and compute all possible $d_{\mathcal{I}}(\bm{s}_m,\bm{s}_{m'})$'s.
Accordingly, the distance between the two event sequences can be defined as the average of the $d_{\mathcal{I}}(\bm{s}_m,\bm{s}_{m'})$'s, i.e.,
\begin{eqnarray}\label{eq:distance}
\begin{aligned}
    d(\bm{s}_m,\bm{s}_{m'}) =\frac{1}{2^{C+1}}\sideset{}{_{\mathcal{I}\in \mathcal{I}_{all}}}\sum d_{\mathcal{I}}(\bm{s}_m,\bm{s}_{m'}),
\end{aligned}
\end{eqnarray}
where $\mathcal{I}_{all}$ represents the set of all possible index subsets.

Given $M$ event sequences, we can compute the distance matrix for them based on~\eqref{eq:distance}, i.e., $\bm{D}=[d(\bm{s}_m,\bm{s}_{m'})]\in\mathbb{R}^{M\times M}$. 
Accordingly, we can impose a kernel function on the distance, resulting in another kernel matrix to capture the similarity between arbitrary two sequences. 
Similar to~\eqref{eq:kernel1}, we have
\begin{eqnarray}\label{eq:kernel2}
\begin{aligned}
    \widetilde{K}(\{\bm{s}_m\}_{m=1}^{M}) =[\tilde{\kappa}(\bm{s}_m,\bm{s}_m')]\in\mathbb{R}^{M\times M},
\end{aligned}
\end{eqnarray}
where $\tilde{\kappa}(\bm{s}_m,\bm{s}_m')=\exp(-\frac{d(\bm{s}_m,\bm{s}_{m'})}{2\sigma^2})$.

As shown in~\cite{iwayama2017definition,xu2017thap}, the nonparametric kernel matrix in~\eqref{eq:kernel2} encodes the clustering structure of the event sequences, and applying a spectral clustering algorithm to it can achieve event sequence clustering. 
Therefore, we can leverage the nonparametric kernel matrix to regularize the embedding-based kernel matrix in the training phase, making the sequence-level embeddings inherit the clustering structure. 
Combining the regularization with the MLE framework, we can learn the TPP as follows:
\begin{eqnarray}\label{eq:mle_reg}
\begin{aligned}
    \sideset{}{_{\theta}}\min -\sideset{}{_{m=1}^{M}}\sum\log\mathcal{L}(\bm{s}_m;\theta) + \tau \mathcal{R}(\bm{K}(\theta),\widetilde{\bm{K}}),
\end{aligned}
\end{eqnarray}
where $\mathcal{R}$ denotes the proposed regularizer, which penalizes the discrepancy between the embedding-based kernel and the nonparametric kernel.
The embedding-based kernel is a function of model parameter $\theta$, so we represent it as $\bm{K}(\theta)$.
The hyperparameter $\tau>0$ controls the significance of the regularizer.

% \textcolor{red}{Typically, we can implement the regularizer $\mathcal{R}$ as the mean squared error (MSE) between $\bm{K}(\theta)$ and $\widetilde{\bm{K}}$, i.e., $\|\bm{K}(\theta)-\widetilde{\bm{K}}\|_F^2$. 
% However, such a na\"{i}ve implementation suffers from high computational complexity. 
% Given $M$ event sequences, each of which has $\mathcal{O}(N)$ events and $C$ event types, the computational complexity of the nonparametric kernel matrix is $\mathcal{O}(2^{C+1}N^2M^2)$,\footnote{The computational complexity of the distance in~\eqref{eq:distance} is $\mathcal{O}(2^{C+1}N^2)$, and we need to compute the distance $\mathcal{O}(M^2)$ times for all event sequence pairs.} which is too high to large-scale applications (e.g., a large number of long sequences).} 

\subsubsection{Scalable Implementation}
In theory, we can implement the regularizer $\mathcal{R}$ as the mean squared error (MSE) between $\bm{K}(\theta)$ and $\widetilde{\bm{K}}$, i.e., $\|\bm{K}(\theta)-\widetilde{\bm{K}}\|_F^2$. 
Unfortunately, such a na\"{i}ve implementation is often intractable due to its high computational complexity. 
Given $M$ event sequences, each of which has $\mathcal{O}(N)$ events and $C$ event types, the computational complexity of the nonparametric kernel matrix is $\mathcal{O}(2^{C+1}N^2M^2)$,\footnote{The complexity of the distance in~\eqref{eq:distance} is $\mathcal{O}(2^{C+1}N^2)$, where $2^{C+1}$ is the number of all possible index subsets and $N^2$ means considering the discrepancies for all event pairs. We need to compute $\mathcal{O}(M^2)$ distances for all event sequence pairs.} which is too high to large-scale applications (e.g., modeling a large number of long sequences). 
To derive a scalable regularizer, we propose an efficient implementation with the help of random sampling and optimal transport techniques~\cite{memoli2011spectral,peyre2016gromov}.

Firstly, when computing the nonparametric distance $d(\bm{s}_m,\bm{s}_{m'})$, instead of enumerating all $2^{C+1}$ possible index subsets, we only consider $C+1$ subsets, each of which contains a single index. 
Therefore, for arbitrary two event sequences, we approximate their distance with the computational complexity $\mathcal{O}(CN^2)$. 
The work in~\cite{iwayama2017definition} has shown that the distance based on the sampled subsets can preserve strong discriminative power.
Secondly, instead of computing a full-sized nonparametric kernel matrix, we sample $L$ event sequence randomly from the dataset $\mathcal{S}$ and construct a small kernel matrix, i.e., $\widehat{\bm{K}}_L\in\mathbb{R}^{L\times L}$ and $L\ll M$.
The computational complexity of $\widehat{\bm{K}}_L$ is $\mathcal{O}(CN^2L^2)$, which is much lower than that of $\widetilde{\bm{K}}$. 

The sampling of event sequences breaks the one-one correspondence between the embedding-based kernel matrix and the nonparametric kernel matrix, making the MSE loss inapplicable. 
In this study, we leverage the Gromov-Wasserstein (GW) distance~\cite{memoli2011gromov} as a surrogate, measuring the discrepancy between the embedding-based kernel matrix and the approximated nonparametric kernel matrix. 
The GW distance provides a valid metric for metric-measure spaces~\cite{memoli2011gromov}, which can be extended to measure the distance between two kernel functions~\cite{memoli2011spectral}. 
Denote $\mathcal{X}_{\mu,\kappa_1}$ and $\mathcal{Y}_{\nu,\kappa_2}$ as two metric-measure spaces, respectively, where $\mu$, and $\nu$ are probability measures on the two spaces, and $\kappa_1:\mathcal{X}^2\mapsto \mathbb{R}_+$ and $\kappa_2:\mathcal{Y}^2\mapsto \mathbb{R}_+$ are two kernel functions defined in the two spaces.
The $p$-order GW distance between the two kernel functions is defined as
% \begin{eqnarray}\label{eq:gwd}
% \begin{aligned}
%     &{GW}_p(\kappa_1,\kappa_2)=\inf_{\pi\in\Pi_{\mu,\nu}}\mathbb{E}_{x,y,x',y'\sim\pi\times\pi}^{1/p}[|\kappa_1(x,x')-\kappa_2(y,y')|^p]\\
% &=\sideset{}{_{\pi\in\Pi_{\mu,\nu}}}\inf\Bigl(\int_{\mathcal{X}^2\times\mathcal{Y}^2}r^p(x,x',y,y')\mathrm{d}\pi(x,y)\mathrm{d}\pi(x',y')\Bigr)^{\frac{1}{p}},
% \end{aligned}
% \end{eqnarray}
\begin{eqnarray}\label{eq:gwd}
\begin{aligned}
    &{GW}_p(\kappa_1,\kappa_2)\\
    &= \sideset{}{_{\pi\in\Pi_{\mu,\nu}}}\inf \mathbb{E}_{x,y,x',y'\sim\pi\times\pi}^{1/p} \bigl[|\kappa_1(x,x')-\kappa_2(y,y')|^p\bigr] \\
    &= \sideset{}{_{\pi\in\Pi_{\mu,\nu}}}\inf 
    \Bigl(\int_{\mathcal{X}^2\times\mathcal{Y}^2} r^p(x,x',y,y') \, \mathrm{d}\pi(x,y) \, \mathrm{d}\pi(x',y')\Bigr)^{\frac{1}{p}},
\end{aligned}
\end{eqnarray}

where $\pi$ is called transport plan or coupling between $\mu$ and $\nu$. 
It is a distribution defined on $\mathcal{X}\times\mathcal{Y}$, whose marginals are $\mu$ and $\nu$, respectively, i.e., $\pi\in\Pi_{\mu,\nu}=\{\pi\geq 0|\int_{\mathcal{X}}\mathrm{d}\pi(x,y)=\nu(y),\int_{\mathcal{Y}}\mathrm{d}\pi(x,y)=\mu(x)\}$.
$r(x,x',y,y')=|\kappa_1(x,x')-\kappa_2(y,y')|$ is called ``relational distance''~\cite{xu2019scalable}, which measures the distance between the two kernel functions given two sample pairs (i.e. $(x,x')$ and $(y,y')$).
As shown in~\eqref{eq:gwd}, the GW distance corresponds to the infimum of the expected relational distance. 
The transport plan corresponding to the infimum is called the optimal transport plan, denoted as $\pi^*$.

Given two kernel matrices sampled from the two kernel functions, i.e., $\bm{K}_1\in\mathbb{R}^{M\times M}$ and $\bm{K}_2\in\mathbb{R}^{L\times L}$, we can define the empirical $p$-order GW distance accordingly. 
When $p=2$, the empirical GW distance leads to a constrained quadratic optimization problem~\cite{peyre2016gromov}:
\begin{eqnarray}\label{eq:gwd2}
\begin{aligned}
    &\widehat{GW}_2(\bm{K}_1,\bm{K}_2) = \sideset{}{_{\bm{T}\in\Pi_{\bm{\mu},\bm{\nu}}}}\min \langle\bm{C}(\bm{K}_1,\bm{K}_2,\bm{T}), \bm{T}\rangle^{1/2} \\
    &= \sideset{}{_{\bm{T}\in\Pi_{\bm{\mu},\bm{\nu}}}}\min 
    \mathbb{E}_{m,l,m',l'\sim\bm{T}\times\bm{T}}^{1/2} \bigl[|\bm{K}_1(m,m')-\bm{K}_2(l,l')|^2\bigr],
\end{aligned}
\end{eqnarray}
where $\bm{C}(\bm{K}_1,\bm{K}_2,\bm{T})=(\bm{K}_1\odot\bm{K}_1)\bm{\mu}\bm{1}_L^{\top} +\bm{1}_M\bm{\nu}^{\top}(\bm{K}_2\odot\bm{K}_2) - 2\bm{K}_1\bm{T}\bm{K}_2^{\top}$, $\odot$ is the Hadamard product, $\bm{\mu}=\frac{1}{M}\bm{1}_M$ and $\bm{\nu}=\frac{1}{L}\bm{1}_L$ are two empirical distributions, and $\bm{T}=[t_{ml}]$ is the transport matrix taking $\bm{\mu}$ and $\bm{\nu}$ as its marginals.
$\Pi_{\bm{\mu},\bm{\nu}}=\{\bm{T}\in\mathbb{R}_{+}^{M\times L}|\bm{T1}_{L}=\bm{\mu},\bm{T}^{\top}\bm{1}_M=\bm{\nu}\}$ is the feasible domain of $\bm{T}$.
The problem in~\eqref{eq:gwd2} can be solved iteratively by the proximal gradient algorithm~\cite{xu2019scalable}. 
The complexity of the algorithm is $\mathcal{O}(M^2L+L^2M)$, and its convergence is guaranteed in theory --- with the increase of the iterations, $\bm{T}$ converges to a stationary point. 

The above empirical GW distance measures the discrepancy between arbitrary two kernel matrices, in which the optimal transport matrix, denoted as $\bm{T}^*$, indicates the correspondence between the two matrices' rows/columns. 
It has been shown in~\cite{xu2019scalable,chowdhury2021generalized} that when one matrix reflects the data similarity while the other matrix encodes the clustering structure (e.g., a diagonally dominant matrix), computing the empirical GW distance achieves the clustering of the data, in which $\bm{T}^*$ reflects the coherency probability of each data point and each cluster, i.e., $t_{ml}^*$ is the probability that the $m$-th data point belongs to the $l$-th cluster.
Therefore, we implement our clustering regularizer as the empirical GW distance between $\bm{K}(\theta)$ and $\widehat{\bm{K}}_L$, i.e., $\mathcal{R}(\bm{K}(\theta), \widehat{\bm{K}}_L)=\widehat{GW}_2^2(\bm{K}(\theta), \widehat{\bm{K}}_L)$.
As a result, our learning problem becomes
% \begin{eqnarray}\label{eq:obj2}
% \begin{aligned}
%     \sideset{}{_{\theta}}\min -\sideset{}{_{m=1}^{M}}\sum\log\mathcal{L}(\bm{s}_m;\theta) + \tau \widehat{GW}_2^2(\bm{K}(\theta),\widehat{\bm{K}}_L). 
% \end{aligned}
% \end{eqnarray}
\begin{eqnarray}\label{eq:obj2}
\begin{aligned}
    \sideset{}{_{\theta}}\min -\sideset{}{_{m=1}^{M}}\sum \log\mathcal{L}(\bm{s}_m; \theta) 
    + \tau \widehat{GW}_2^2(\bm{K}(\theta), \widehat{\bm{K}}_L).
\end{aligned}
\end{eqnarray}

This problem can be solved efficiently by mini-batch stochastic gradient descent (SGD). 
Given a batch of event sequences, we apply an alternating optimization strategy to compute the optimal transport matrix and update the model parameter. 

\section{Experiments}
To demonstrate the effectiveness of our method, we evaluate it on several synthetic and real-world datasets.
These experiments highlight the superiority of our method, and further analytic studies are conducted to analyze the interpretability and scalability of the method.
All experiments are run on a server with two Nvidia 3090 GPUs.

% (段落和章节之间要空行！要不然别人polish不方便)
\subsection{Implementation Details}
\subsubsection{Datasets}
% The synthetic datasets provided by ~\cite{zhang2022learning} generate event sequences with 5 different processes, which include the Homogeneous Poisson process, 
% In-homogeneous Poisson process, Inhibit Process, Excite Process (Hawkes) and In\&Ex Process in which relation is either inhibition or excitation.
% Each process generates 4,000 event sequences, and each sequence contains 50 events with the number of event types $C = 5$.
We conducted experiments using both synthetic datasets and real-world datasets. 
\begin{itemize}
    \item \textbf{Synthetic dataset}~\cite{zhang2022learning} consists of the event sequences generated by four different TPPs (i.e., In-homogeneous Poisson process, Inhibit process, Hawkes process, and In\&Ex Process in which the event relation is either inhibition or excitation).
    Each TPP generates 4,000 event sequences with $C=5$ event types, and each sequence contains 50 events.
    \item \textbf{Taobao}~\cite{xue2022hypro} consists of the time-stamped browsing behavior sequences of the 2,000 most active users on an online shopping platform called Taobao.
    The events are categorized into $C=17$ event types corresponding to item classes (e.g., men's clothes).
    \item \textbf{StackOverflow (SO)}~\cite{linderman2014discovering} contains 2,200 user award sequences on a question-answering website: each user received a sequence of badges, and there are $C=22$ different kinds of badges in total. 
    Each sequence represents a user’s reward history and each reward (i.e., event) contains a timestamp and a badge (i.e., event type).
    \item \textbf{Retweet}~\cite{zhao2015seismic} contains 24,000 user behavior sequences collected from Twitter.
    Each sequence consists of time-stamped tweets, and each tweet is treated as an event, which is categorized into $C=3$ event types based on the number of the user's followers.
    \item \textbf{Taxi}~\cite{whong2014foiling} captures the time-stamped taxi pick-up and drop-off events across the five boroughs of New York City. 
    Each unique combination of a borough and a pick-up or drop-off event constitutes a distinct event type, resulting in a total of $C=10$ event types. 
    The dataset comprises 2,000 drivers' event sequences.
\end{itemize}

% The basic statistics of the datasets are shown in Table~\ref{tab:data}.

% \begin{table}[t]
% \centering
% \caption{The basic statistics of datasets.}\label{tab:data}
% % \small{
% \begin{tabular}{c|ccc}
%   \hline
%   \hline
%   Dataset & $C$ & \#Sequences & Max length \\
%   \hline
%   Synthetic ($K=2$) & 5 & 8,000 & 50\\
%   Synthetic ($K=3$) & 5 & 12,000 & 50\\
%   Synthetic ($K=4$) & 5 & 16,000 & 50\\
%   \hline
%   \hline
% \end{tabular}
% % }
% \end{table}

\begin{table*}[t]
  \centering
  \caption{Comparison Experiments on Synthetic Datasets}
  \small{
  \tabcolsep=2pt
    \begin{tabular}{c|cccc|cccc|cccc}
    \toprule
    \multirow{3}[3]{*}{Method} & 
    % \multicolumn{4}{c|}{$K_{GT}=2$} & 
    \multicolumn{4}{c|}{Synthetic ($K=2$)} & 
    \multicolumn{4}{c|}{Synthetic ($K=3$)} & 
    \multicolumn{4}{c }{Synthetic ($K=4$)} \\
    \cmidrule{2-13}
    &
    % \multicolumn{2}{c}{Pred} & 
    % \multicolumn{2}{c|}{Cluster} & 
    \multicolumn{2}{c}{Prediction} & 
    \multicolumn{2}{c|}{Clustering} & 
    \multicolumn{2}{c}{Prediction} & 
    \multicolumn{2}{c|}{Clustering} & 
    \multicolumn{2}{c}{Prediction} & 
    \multicolumn{2}{c}{Clustering} \\
    \cmidrule{2-13} 
    % \cmidrule{15-18}
    & 
    ELL$\uparrow$ & Acc$\uparrow$ & NMI$\uparrow$ & RI$\uparrow$ & 
    ELL$\uparrow$ & Acc$\uparrow$ & NMI$\uparrow$ & RI$\uparrow$ & 
    ELL$\uparrow$ & Acc$\uparrow$ & NMI$\uparrow$ & RI$\uparrow$  \\
    \midrule
    DIS+SC & 
    - & - & 0.788 & 0.869 & 
    - & - & 0.866 & 0.898 & 
    - & - & 0.583 & 0.531  \\
   %- & - & 0.509 & 0.432 
    \midrule
    {RMTPP} & 
    -0.518$_{0.003}$ & 0.291$_{0.002}$ & 0.732$_{0.008}$ &  0.819$_{0.008}$ & 
    -0.673$_{0.005}$ & 0.285$_{0.003}$ & 0.787$_{0.007}$ & 0.834$_{0.008}$ & 
    \textbf{-0.747}$_{0.001}$ & 0.261$_{0.003}$ & 0.551$_{0.012}$ & 0.491$_{0.019}$ \\
    % \cmidrule(lr){15-18}
    RMTPP+$\mathcal Reg$ & 
    \textbf{-0.517}$_{0.002}$ & \textbf{0.292}$_{0.000}$ & \textbf{0.754}$_{0.030}$ & \textbf{0.840}$_{0.025}$ & 
    \textbf{-0.673}$_{0.004}$ & \textbf{0.285}$_{0.002}$ & \textbf{0.795}$_{0.016}$ & \textbf{0.842}$_{0.019}$ & 
    -0.747$_{0.002}$ & \textbf{0.262}$_{0.003}$ & \textbf{0.575}$_{0.020}$ & \textbf{0.529}$_{0.055}$ \\ 
    \midrule
    {NHP} & 
    -0.456$_{0.002}$ & 0.295$_{0.002}$ & 0.707$_{0.014}$ & 0.772$_{0.012}$ & 
    \textbf{-0.585}$_{0.003}$ & 0.285$_{0.004}$ & 0.882$_{0.002}$ & 0.913$_{0.001}$ & 
    -0.636$_{0.004}$ & 0.272$_{0.002}$ & 0.803$_{0.008}$ & 0.801$_{0.009}$ \\ 
    % \cmidrule(lr){15-18}
    NHP+$\mathcal Reg$ & 
    \textbf{-0.452}$_{0.000}$ & \textbf{0.295}$_{0.000}$ & \textbf{0.815}$_{0.043}$ & \textbf{0.883}$_{0.049}$ & 
    -0.585$_{0.004}$ & \textbf{0.286}$_{0.007}$ & \textbf{0.887}$_{0.003}$ & \textbf{0.919}$_{0.006}$ & 
    \textbf{-0.636}$_{0.003}$ & \textbf{0.272}$_{0.001}$ & \textbf{0.807}$_{0.003}$ & \textbf{0.807}$_{0.006}$ \\ 
    \midrule
    {THP} & 
    1.053$_{0.006}$ & 0.248$_{0.005}$ & 0.661$_{0.057}$ & 0.750$_{0.055}$ & 
    0.907$_{0.012}$ & 0.273$_{0.009}$ & 0.397$_{0.212}$ & 0.288$_{0.280}$ & 
    0.940$_{0.000}$ & 0.244$_{0.003}$ & 0.120$_{0.022}$ & 0.034$_{0.022}$ \\ 
    % \cmidrule(lr){15-18}
    THP+$\mathcal Reg$ & 
    \textbf{1.053}$_{0.005}$ & \textbf{0.251}$_{0.005}$ & \textbf{0.742}$_{0.033}$ & \textbf{0.831}$_{0.027}$ & 
    \textbf{0.907}$_{0.011}$ & \textbf{0.273}$_{0.009}$ & \textbf{0.476}$_{0.102}$ & \textbf{0.397}$_{0.122}$ & 
    \textbf{0.940}$_{0.000}$ & \textbf{0.246}$_{0.003}$ & \textbf{0.344}$_{0.036}$ & \textbf{0.205}$_{0.015}$ \\ 
    \bottomrule
    \end{tabular}%
  }
  \label{tab:cmp}%
\end{table*}%

\subsubsection{Baselines and Backbone Models}
In this study, we consider both parametric TPP models and nonparametric ones.
We take the work in~\cite{iwayama2017definition} as a baseline, which computes the nonparametric distance matrix for event sequences and then applies spectral clustering (\textbf{DIS+SC}).
Additionally, the state-of-the-art mixture model of neural TPPs in~\cite{zhang2022learning} is considered in the experiments on real-world datasets.
For our method, we consider the following three models as backbones, demonstrating the universality of the proposed regularizer.
\begin{itemize}
    \item \textbf{RMTPP}~\cite{du2016recurrent} and \textbf{NHP}~\cite{mei2017neural} are neural TPPs that leverage recurrent neural networks in the continuous-time domain.
    \item \textbf{THP}~\cite{zuo2020transformer} is one of the state-of-the-art neural TPPs that applies a Transformer-like architecture.
\end{itemize}
For our method, the bandwidth $\sigma$ of kernel matrix is a key hyperparameter. 
For the nonparametric kernel matrix, we apply an adaptive method to determine the bandwidth.
In particular, given the distances among the event sequences, we empirically set $\sigma$ based on the median of the distances.
For the remaining hyperparameters, e.g., learning rate, batch size, epochs, and so on, we configure them based on the default settings in~\cite{xue2024easytpp} for a fair comparison. 
We train the above backbone models in the MLE framework. 
The models trained purely based on the MLE and those trained by the MLE with our regularizer are compared on the following evaluation measurements.

% \xu{Add a comparison for Mixtured TPP and our model on their model size and inference time. Ideally, our model size should be much smaller and the inference time should be faster.}

\subsubsection{Evaluation Measurements} 
Given a learned model, we use $i)$ the log-likelihood per event (\textbf{ELL}) and $ii)$ the prediction accuracy of event types (\textbf{ACC}) to evaluate its data fidelity and prediction power, respectively. 
When the model is learned on synthetic datasets, whose event sequences are associated with cluster labels, we employ $i)$ Normalized Mutual Information (\textbf{NMI}) and $ii)$ Rand Index (\textbf{RI}) to evaluate the model's clustering performance.

\subsection{Event Sequence Clustering and Prediction} 
\subsubsection{Comparisons on Synthetic Data}
A comprehensive set of comparison experiments are conducted to assess the performance of our proposed method against the baselines. 
The results of these experiments are summarized in Table~\ref{tab:cmp}, showcasing the superior performance of our method across different datasets and evaluation metrics. 
In particular, the classic nonparametric clustering method~\cite{iwayama2017definition} computes the distance matrix for event sequences and then applies spectral clustering. 
This method is only applicable for clustering tasks and its performance degrades a lot concerning the number of clusters.
For parametric TPP models, i.e., RMTPP, NHP, and THP, the original MLE-based learning paradigm does not impose any constraint on their sequence-level embeddings, so the clustering power of the learned embeddings is limited.
After applying the proposed regularizer, we can find that these models have improvements on clustering tasks, while their prediction power is preserved well or improved simultaneously.

\begin{figure}[t]
    \centering
    \subfigure[THP]{
    \includegraphics[height=2.7cm]{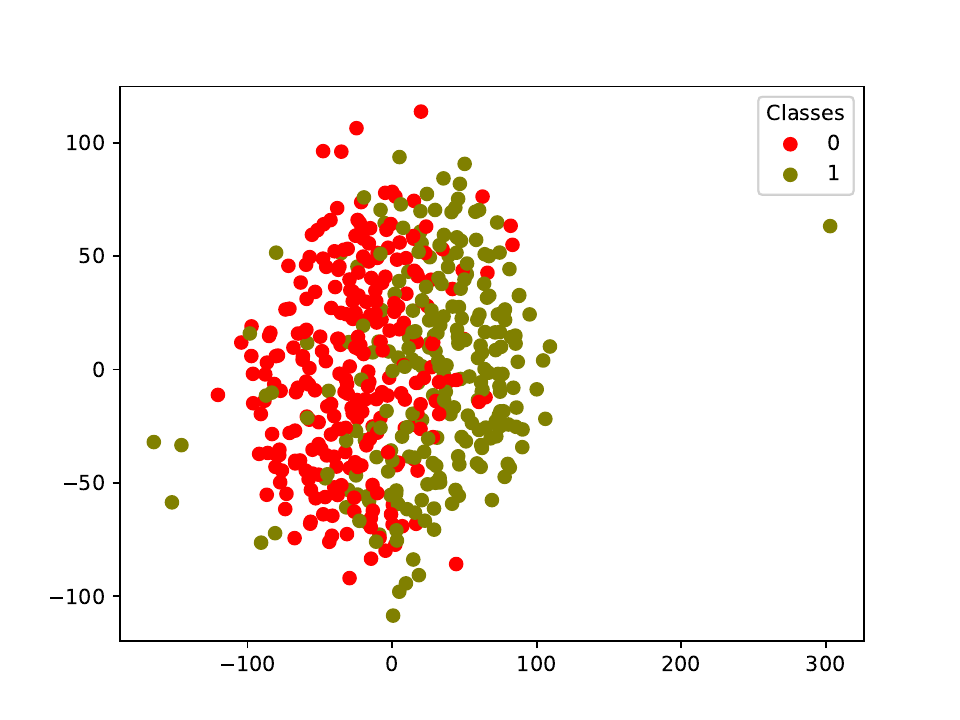}\label{fig:tsne1}
    }
    \subfigure[THP+$\mathcal Reg$]{
    \includegraphics[height=2.7cm]{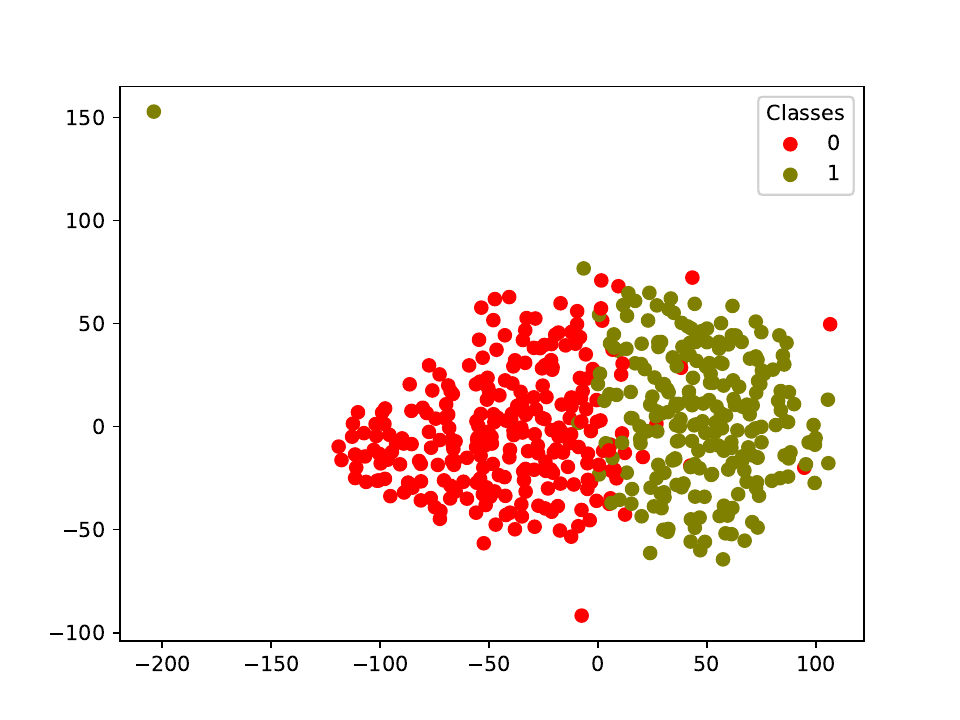}\label{fig:tsne2}
    }\\

    \subfigure[Nonparametric $\widehat{\bm{K}}$]{
    \includegraphics[height=2.5cm]{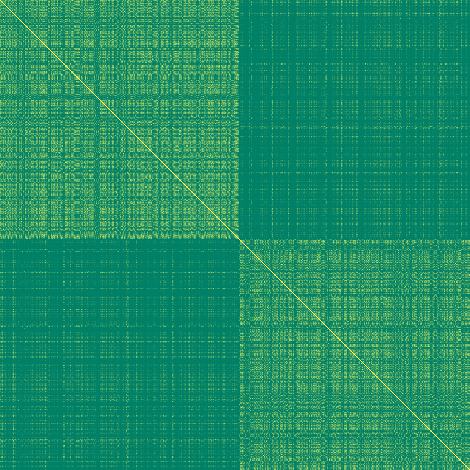}\label{fig:kernel1}
    }
    \subfigure[$\bm{K}(\theta)$ of THP]{
    \includegraphics[height=2.5cm]{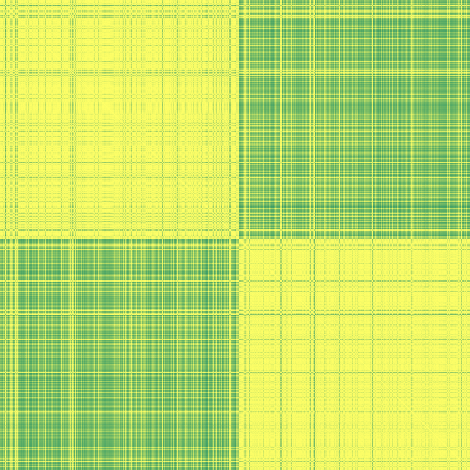}\label{fig:kernel2}
    }
    \subfigure[$\bm{K}(\theta)$ of THP+$\mathcal Reg$]{
    \includegraphics[height=2.5cm]{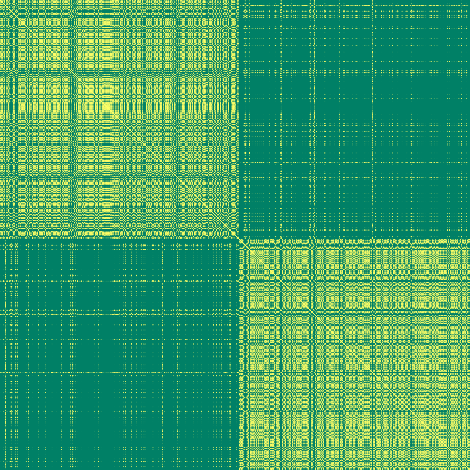}\label{fig:kernel3}
    }
    \caption{An illustration of the improvement on clustering caused by our regularizer. 
    The backbone model is THP~\cite{zuo2020transformer} and the event sequences are from the synthetic dataset ($K=2$). 
    In (a, b), we sample 500 event sequences per cluster and visualize their embeddings by t-SNE.
    Furthermore, we visualize the kernel matrices obtained by (c) the nonparametric method in~\cite{iwayama2017definition}, (d) the embedding-based kernel obtained by the original THP model, and (e) the embedding-based kernel obtained by the THP learned with our regularizer.
    % \xu{Add more t-SNE plots, for more backbones and more datasets.}
    }
    \label{fig:visual}
\end{figure}

% \begin{figure*}[t]
%     \centering
%     \subfigure[NSM]{
%     \includegraphics[height=4.5cm]{WWW24/figures/test_NSM_kgt2.png}
%     }
%     \subfigure[THP]{
%     \includegraphics[height=4.5cm]{WWW24/figures/test_THP_without_reg_kgt2.png}
%     }
%     \subfigure[THP+Reg]{
%     \includegraphics[height=4.5cm]{WWW24/figures/test_THP_reg_kgt2.png}
%     }
%     \caption{The heatmap.}
%     \label{fig:robust}
% \end{figure*}

\subsubsection{Visualization of Clustering Results}
% To further demonstrate the rationality of our method, we shows the t-SNE plots of the learned sequence-level embeddings in Figure~\ref{fig:visual}.
Given the embeddings obtained by the original THP model and those achieved by the model trained with our regularizer, we show their t-SNE plots in Figure~\ref{fig:tsne1} and~\ref{fig:tsne2}, respectively.
According to the visual effects, we can find that the embeddings learned by our method have more distinguishable clustering structures, i.e., the sequence embeddings corresponding to different classes are separated while those within the same class are concentrated. 
On the contrary, without our regularizer, the embeddings of the original THP tends are mixed and do not have significant clustering structures.

Besides the t-SNE plots, the kernels constructed by the embeddings also demonstrate the superiority of our method.
As shown in Figures~\ref{fig:kernel1}-\ref{fig:kernel3}, we visualize the kernel matrix constructed by the nonparametric method in~\cite{iwayama2017definition}, the embedding-based kernel obtained by the original THP, and that obtained by our THP+Reg method. 
We can find that the kernel obtained by our method has a more significant blockwise structure than the remaining two kernels, which results in better clustering results. 

In our opinion, this phenomenon can be explained as follows.
Essentially, both the nonparametric kernel $\widehat{\bm{K}}$ and the $\bm{K}(\theta)$ of the original THP are highly-noisy due to the randomness of event happenings in the sequences. 
As a result, the clustering results based on such kernels are often unsatisfactory, as shown in Table~\ref{tab:cmp}.
Our method provides an effective framework considering these two kernels jointly.
Through the proposed regularizer, these two kernels provide useful prior information with each other and thus are mutually reinforced during training, leading to a kernel with better clustering structures.

\begin{table*}[t]
  \centering
  \caption{Comparison Experiments on Real-world Datasets.}
  \small{
  \tabcolsep=2pt
    \begin{tabular}{c|ccc|ccc|ccc|ccc}
    \toprule
    \multirow{2}{*}{Method} & \multicolumn{3}{c|}{Taobao} & \multicolumn{3}{c|}{StackOverflow} & \multicolumn{3}{c|}{Retweet}  & \multicolumn{3}{c}{Taxi}\\
    & ELL $\uparrow$  & ACC $\uparrow$ & \#Param $\downarrow$
    & ELL $\uparrow$  & ACC $\uparrow$ & \#Param $\downarrow$
    & ELL $\uparrow$  & ACC $\uparrow$ & \#Param $\downarrow$
    & ELL $\uparrow$  & ACC $\uparrow$ & \#Param $\downarrow$\\
    \midrule
    % RMTPP 
    % &&&
    % &&&
    % &&&
    % &&&\\
    Mixed RMTPPs
    & \textbf{-0.214}$_{0.002}$  & 0.252$_{0.001}$  &10,621
    & \textbf{-2.707}$_{0.001}$  & 0.425$_{0.000}$  &11,786
    & \textbf{-4.086}$_{0.001}$  & 0.561$_{0.007}$  &7,359
    & \textbf{0.304}$_{0.001}$  &0.905$_{0.002}$  & 8,990\\
    RMTPP+$\mathcal Reg$ 
    &-0.484$_{0.010}$ &\textbf{0.436}$_{0.000}$ & \textbf{3,347}
    &-2.749$_{0.003}$ & \textbf{0.425}$_{0.000}$& \textbf{3,682}
    &-4.108$_{0.026}$ & \textbf{0.565}$_{0.007}$& \textbf{2,409}
    &0.271$_{0.008}$ &\textbf{0.909}$_{0.001}$& \textbf{2,878}\\
    \midrule
    % NHP 
    % &&&
    % &&&
    % &&&
    % &&&\\
    Mixed NHPs 
    & 0.729$_{0.016}$ & 0.509$_{0.001}$ &181,252 
    & \textbf{-2.285}$_{0.024}$   & 0.450$_{0.001}$&183,492
    & \textbf{-3.568}$_{0.031}$   &0.566$_{0.006}$ &181,252
    &\textbf{0.516}$_{0.001}$ &0.896$_{0.000}$& 178,116\\
    NHP+$\mathcal Reg$ 
    & \textbf{0.893}$_{0.001}$ &\textbf{0.602}$_{0.000}$  &\textbf{60,032}
    & -2.460$_{0.028}$  &\textbf{0.452}$_{0.001}$ &\textbf{60,672}
    & -3.809$_{0.027}$    & \textbf{0.620}$_{0.002}$&\textbf{58,240}
    &0.484$_{0.012}$ &\textbf{0.897}$_{0.000}$&\textbf{59,136}\\
    \midrule
%     Mixed SAHPs 
%     & \textbf{-0.514}$_{0.020}$ &0.238$_{0.040}$ &23,431
%     & \textbf{-13.524}$_{0.626}$  & 0.020$_{0.004}$ &24,374
% &-94.354$_{164.188}$
% &0.447$_{0.005}$&39069
%     &\textbf{0.762}$_{0.060}$ & 0.740$_{0.036}$& 21,842\\
%     SAHP+$\mathcal Reg$ &-0.618
% $_{0.081}$ &\textbf{0.427}$_{0.007}$ &\textbf{7,489}
%     &-13.911$_{0.200}$ &\textbf{0.037}$_{0.006}$&\textbf{7814}
%     &\textbf{-23.910}$_{1.346}$ &\textbf{0.448}$_{0.044}$ &\textbf{6579}
%     &0.504$_{0.026}$ 
%     &\textbf{0.892}$_{0.010}$& \textbf{7034}\\
%     \midrule
%     % THP 
%     % &&&
%     % &&&
%     % &&&
%     % &&&\\
    Mixed THPs 
    & \textbf{-0.127}$_{0.028}$ & 0.449$_{0.001}$ &22,999
    &\textbf{-2.402}$_{0.005}$   &0.450$_{0.002}$  & 24134
    & \textbf{-4.636}$_{0.230}$    & 0.562$_{0.002}$ &19,821
    & \textbf{0.244}$_{0.021}$ &0.909$_{0.001}$& 21,410\\
    THP+$\mathcal Reg$ 
    &-0.419$_{0.013}$ &\textbf{0.475}$_{0.009}$ &\textbf{7,473}
    &-2.433$_{0.008}$  &\textbf{0.453}$_{0.001}$  &\textbf{7,798}
    &-5.148$_{0.363}$ &\textbf{0.580}$_{0.033}$ &\textbf{6,563}
    & 0.234$_{0.007}$ 
    & \textbf{0.911}$_{0.001}$& \textbf{7018}\\
    \bottomrule
  \end{tabular}
  }
  \label{tab:real}
\end{table*}

\subsubsection{Comparisons on Real-world Data}
Besides the above synthetic experiments, we test our method on four real-world datasets and compare it with the mixture model of TPPs~\cite{zhang2022learning}.
The backbone TPPs used in the mixture model and our method include RMTPP, NHP and THP.
Because the real-world data do not have clustering labels, we mainly focus on the performance of the models on their data fitness (i.e., testing log-likelihood) and event prediction power (i.e., prediction accuracy). 
In addition, to highlight the scalability of our method, the number of parameters for each learned model is recorded as well. 
Table~\ref{tab:real} shows the experimental results.
We can find that our method outperforms the mixture model consistently on the prediction accuracy while degrades slightly on the testing log-likelihood. 
Because of learning a single TPP, our method reduces the number of parameters significantly compared to learning mixed TPPs. 
A potential reason for this phenomenon is that the mixture model leverage multiple TPPs to fit different clusters of event sequences, which can fit data better than a single TPP does in general. 
However, when predicting future events, it has to first determine the cluster of each testing sequence and then make predictions based on the selected TPP component, which may suffer the error propagation issue --- the wrongly selected TPP often leads to catastrophic prediction results. 
On the contrary, our regularization approach provides an effective alternative to complex mixture models in predictive tasks, which learns a single TPP to predict future events directly. 
Considering the improvements on prediction accuracy and the reduction of model parameters, a single TPP model learned with our regularizer can still be competitive to the mixed TPPs.

\section{Conclusion}
In this paper, we introduced a novel approach for enhancing the clustering structures of event sequence embeddings for both parameterized and nonparametric TPPs.
We utilized the Gromov-Wasserstein distance to quantify the discrepancy between the parametric kernel derived by sequence embeddings and a sampled nonparametric kernel, subsequently incorporating this as a regularization term in the MLE framework of TPP. 
This enabled us to learn a robust and interpretable parametric TPP efficiently, with enhanced clustering power and competitive prediction performance.

\section*{Acknowledgments} 
This work was supported by National Natural Science Foundation (92270110), the Fundamental Research Funds for the Central Universities, and the Research Funds of Renmin University of China. 
We also acknowledge the support provided by the fund for building world-class universities (disciplines) of Renmin University of China and by the funds from Engineering Research Center of Next-Generation Intelligent Search and Recommendation, Ministry of Education, and from Intelligent Social Governance Interdisciplinary Platform, Major Innovation \& Planning Interdisciplinary Platform for the ``Double-First Class'' Initiative, Renmin University of China.

\bibliographystyle{ACM-Reference-Format}
\bibliography{main}

\end{document}